\documentclass[sn-mathphys, Numbered, iicol]{sn-jnl}

\usepackage{amsmath,amsfonts,amssymb}
\usepackage{array}
\usepackage{amsthm}
\usepackage{textcomp}
\usepackage{stfloats}
\usepackage{url}
\usepackage{verbatim}
\usepackage{graphicx}
\usepackage{multirow}
\usepackage{tensor}
\usepackage{siunitx}
\usepackage{mathrsfs}
\usepackage{booktabs}
\usepackage{color}
\usepackage{pifont}
\usepackage{subcaption} 
\usepackage{algorithm}  
\usepackage{algpseudocode} 

\usepackage[title]{appendix}%
\usepackage{manyfoot}%
\usepackage{algorithmicx}%
\usepackage{listings}%

\usepackage{xcolor}
\definecolor{myOrange}{rgb}{1, 0.5, 0.055} 
\definecolor{myRed}{rgb}{0.84, 0.15, 0.157} 
\definecolor{myGreen}{rgb}{0.173, 0.627, 0.173} %

\newcommand{\cmark}{\ding{51}}
\newcommand{\xmark}{\ding{55}}
\newcommand{\halfcmark}{\checkmark\kern-1.1ex\raisebox{.7ex}{\rotatebox[origin=c]{125}{--}}}

\def\secref#1{Sec.~\ref{#1}}
\def\figref#1{Fig.~\ref{#1}}
\def\tabref#1{Tab.~\ref{#1}}
\def\eqref#1{Eq.~(\ref{#1})}

\def\secref#1{Sec.~\ref{#1}}

\def\figref#1{Fig.~\ref{#1}}
\def\tabref#1{Tab.~\ref{#1}}
\def\eqref#1{Eq.~(\ref{#1})}

\newcommand{\nmat}[1]{\mathbf{#1}} 

\newcommand{\rotation}[2]{\nmat{R}_{#1 #2}} 

\newcommand{\SOthree}{\textrm{SO}(3)}

\newcommand{\SEthree}{\textrm{SE}(3)}

\newcommand{\subarrow}[1]{
	\mathord{
		\renewcommand{\arraystretch}{0}
		\begin{array}[t]{@{}c@{}l@{}}
			#1\\[2pt]
			\hspace{-2pt}\scriptstyle\longrightarrow
		\end{array}
		\kern\scriptspace
	}
}

\newcommand{\legacyframe}[1]{\subarrow{\mathcal{F}}{}_{#1}}
\robustify{\legacyframe}

\usepackage{color}
\definecolor{ColorLightCyan}{rgb}{0.88,1,1}
\definecolor{ColorLightTurquoise}{rgb}{0.5, 1, 0.8}
\definecolor{ColorOrange}{rgb}{1.0, 0.7, 0.0}
\definecolor{ColorTrajBlue}{rgb}{0,0.5,1.0}
\definecolor{ColorTrajOrange}{rgb}{0.87,0.56,0.01}

\usepackage{tikz}

\theoremstyle{thmstyleone}%
\theoremstyle{thmstyletwo}%

\theoremstyle{thmstylethree}%

\begin{document}

\title{Visual Localization in 3D Maps: Comparing Point Cloud, Mesh, and NeRF Representations}

\author*[1]{\fnm{Lintong} \sur{Zhang}}\email{lintong@robots.ox.ac.uk}
\author[1]{\fnm{Yifu} \sur{Tao}}\email{yifu@robots.ox.ac.uk}
\author[2]{\fnm{Jiarong} \sur{Lin}}\email{zivlin@connect.hku.hk}
\author[2]{\fnm{Fu} \sur{Zhang}}\email{fuzhang@hku.hk}
\author[1]{\fnm{Maurice} \sur{Fallon}}\email{mfallon@robots.ox.ac.uk}

\affil*[1]{\orgdiv{Dept. of Engineering Science}, \orgname{Univ. of Oxford}, \orgaddress{\street{Pok Fu Lam}, \city{Hong Kong}, \country{China}}}

\affil[2]{\orgdiv{Dept. of Mechanical Engineering}, \orgname{ Univ. of Hong
Kong}, \orgaddress{\street{Wellington Square}, \city{Oxford}, \country{UK}}}

\abstract{
Recent advances in mapping techniques have enabled the creation of highly accurate dense 3D maps during robotic missions, such as point clouds, meshes, or NeRF-based representations. These developments present new opportunities for reusing these maps for localization. However, there remains a lack of a unified approach that can operate seamlessly across different map representations.
This paper presents and evaluates a global visual localization system capable of localizing a single camera image across various 3D map representations built using both visual and lidar sensing. Our system generates a database by synthesizing novel views of the scene, creating RGB and depth image pairs. Leveraging the precise 3D geometric map, our method automatically defines rendering poses, reducing the number of database images while preserving retrieval performance. To bridge the domain gap between real query camera images and synthetic database images, our approach utilizes learning-based descriptors and feature detectors.
We evaluate the system's performance through extensive real-world experiments conducted in both indoor and outdoor settings, assessing the effectiveness of each map representation and demonstrating its advantages over traditional structure-from-motion (SfM) localization approaches. The results show that all three map representations can achieve consistent localization success rates of \SI{55}{\percent} and higher across various environments. NeRF synthesized images show superior performance, localizing query images at an average success rate of \SI{72}{\percent}. Furthermore, we demonstrate an advantage over SfM-based approaches that our synthesized database enables localization in the reverse travel direction which is unseen during the mapping process. 
Our system, operating in real-time on a mobile laptop equipped with a GPU, achieves a processing rate of \SI{1}{\hertz}.
}

\maketitle

\section{Introduction}
\label{sec:intro}
\begin{figure*}
    \centering
	\includegraphics[width=1\textwidth]{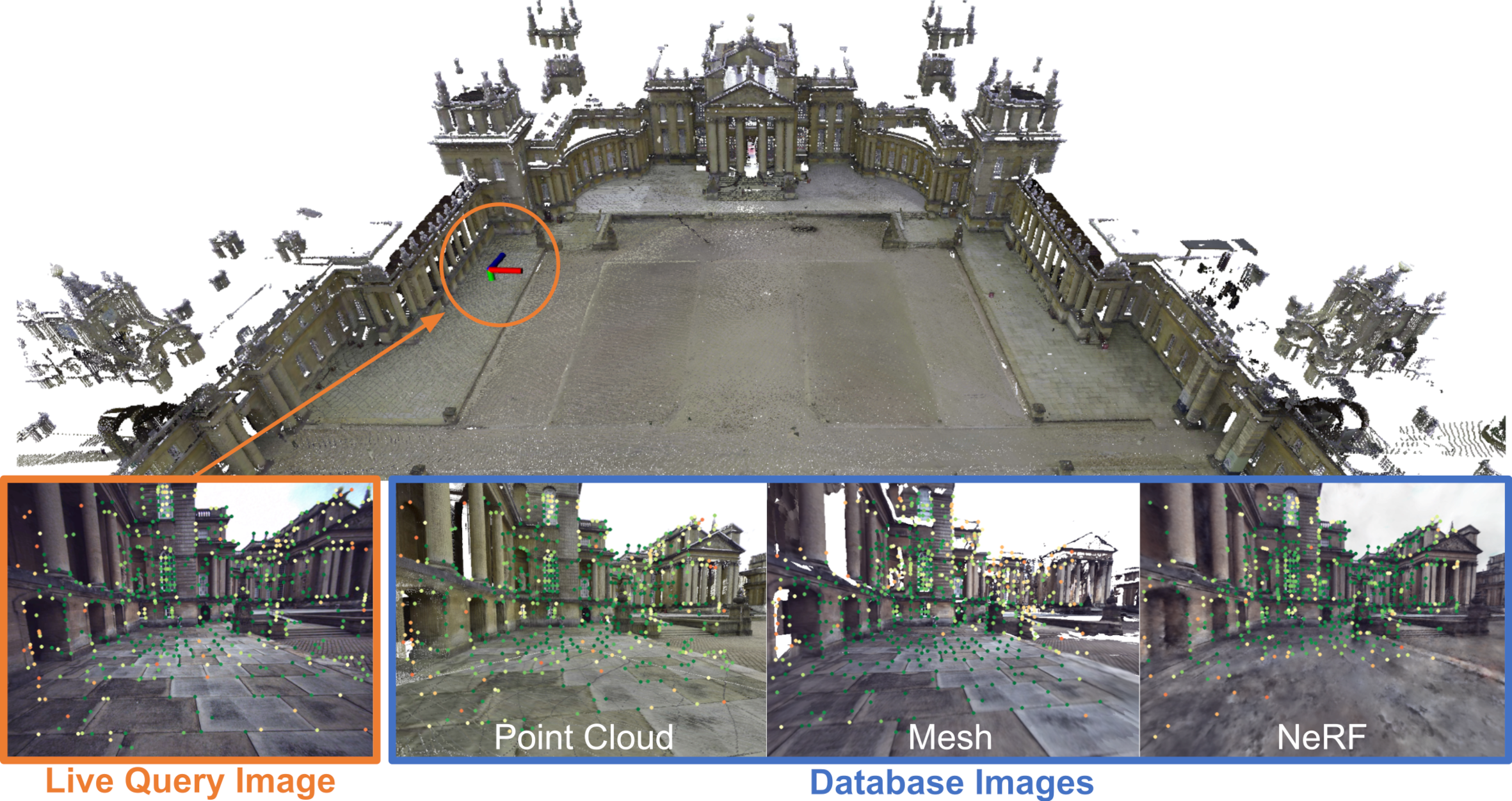}
	\captionof{figure}{Localization of a single query image to a database of images that are synthesized from either point cloud, mesh, or NeRF representations of Blenheim Palace. After identifying a matching image in the database, features are extracted with SuperPoint (as shown above) and matched with SuperGlue. (Displayed point cloud is not directly used.)}
    \label{fig:blenheim_loc}
\end{figure*}%

Recognizing a previously visited place and estimating a robot/sensor's accurate metric pose is fundamental to the problem of robot localization. Localization not only addresses the question of determining where a robot is, but it also enables other tasks such as loop detection in Simultaneous Localization and Mapping (SLAM), multi-robot or multi-session mapping, and augmented reality. Robot localization can be achieved using different sensor modalities, with camera~\cite{sarlin2019coarse, sattler2017, Zeisl2015} and lidar-based approaches ~\cite{scancontext2018, vid2022logg3d, uy2018pointnetvlad} the most commonly studied. While cameras offer a compact form factor and are often low-cost, visual localization and mapping systems often struggle to build large scale accurate 3D maps, especially in real time. Conversely, lidar systems are physically bigger, have higher power consumption and increased cost --- factors which pose challenges for their widespread deployment on mobile robots. Because of this, researchers have proposed combining the advantages of both sensors by constructing a 3D dense map using lidar data (potentially aided by cameras), and then performing localization using only the camera ~\cite{visualLoc2014, visualLocCalib2015}.

For any localization system, a prior representation of the scene---a prior map---is required. The prior map is usually built using the same sensor modality that will be used for later deployment. For example, autonomous industrial inspection has been demonstrated using legged platforms equipped with lidar~\cite{Gehring2021}, where a prior map is built by accumulating laser scans, and localization is performed against the map using of ICP~\cite{Besl1992}. However, lidar localisation is sensitive to physical scene changes and relies on a good initial estimate for ICP registration.
Conversely, purely visual localization methods typically build a map using a structure-from-motion pipeline (SfM) such as COLMAP~\cite{Colmap}, and then localize within it using visual place recognition followed pose refinement using methods such as perspective-n-points (PnP)~\cite{sarlin2019coarse}. SfM maps could, however, be inflexible as they are constrained to the local descriptors used to build them. The pose estimated achieved by visual localization within a SfM map are not scaled to real-world dimensions. As a result, additional information is required for effective deployment in robotic applications. Furthermore, these methods are generally not real-time as the SfM bundle adjustment process can take anywhere from a few hours to an entire day, depending on the size of the images or the map.


Achieving cross-modal localization is typically a more challenging problem due to the inherent differences in sensing modalities and map representations. While lidar-based localization often utilizes point cloud maps, some applications may only have access to a colored mesh-based building model \cite{Yin2023}. In contrast, visual localization methods rely on a database of 3D points with associated descriptors, specifically tailored to the feature detectors used during the map construction \cite{Zeisl2015, sattler2017}. Moreover, emerging scene representations such as NeRF \cite{Liu2023} and Gaussian Splatting \cite{Matsuki2024} are also dependent on the specifics of the original sensing setup. The variety of map representations poses challenges when applying existing localization approaches.

There has been many recent advances in high-quality 3D color mapping --- using either mobile SLAM or fixed TLS data. This has creates new avenues for research into techniques which repurpose these maps for robotic localisation.
We propose to create a unified synthetic map representation/database that accommodates the different types of maps produced by these new mapping systems. In particular, these 3D mapping systems enable the rendering of novel synthetic images from arbitrary rendering positions. In our approach, the visual localisation database is made up of rendered RGB and depth images at chosen poses and can be automatically synthesized from color point clouds, meshes, or NeRF maps. These 3D color maps can be built using vision and lidar data acquired from either industrial-grade Terrestrial Laser Scanners (TLS) or hand-held mapping devices. 
Using this unified representation, we achieve scalable global localization in varied environments, resilient to scene changes and lighting variations with visual learning based components. An overview of the system operation is shown in Fig.~\ref{fig:blenheim_loc}. In addition, we perform a direct comparison with SfM methods to demonstrate the capability of synthesizing views in the opposite direction of travel, as well as highlighting the advantage of reducing the number of required database images. All experiments are conducted in real-world environments, and the corresponding datasets and ground truth maps are available as part of the Oxford Spires Dataset~\footnote{https://dynamic.robots.ox.ac.uk/datasets/oxford-spires/}.

The key contributions of our work include:
\begin{itemize}
\item A versatile and unified approach for global visual localization of a single camera in dense 3D maps, using a database of synthetic RGB and depth images generated from point cloud, mesh, or NeRF representations.
\item A strategy for automatically determining the poses within the color 3D maps from which the synthetic images should be generated.
\item An extensive evaluation across both indoor and outdoor environments comparing visual localisation performance across different map representations.
\item A performance comparison between the our localization system with synthetic images and two (purely) visual localization systems with real images.
\end{itemize}

Our streamlined system operates on a mobile laptop at 1Hz --- and is thus suitable for real-time operation.

\section{Related Work}

We define localization as the process of estimating the 6 DoF pose of a sensor within a 3D prior map or database. This is commonly achieved by means of a place recognition stage---which finds place candidates in the prior map---, and a registration step---that estimates a precise 6 DoF pose using the candidates. 
Given two sensing inputs of vision and lidar, the current literature can be categorized into four categories --- based on sensing modality. These include (a) visual localization within a visual map, (b) lidar localization within a lidar prior map, (c) combined visual and lidar localization within a joint visual and lidar map, and (d) cross-modality visual localization within a lidar prior map or vice versa.

\subsection{Visual Localization}

Visual localization methods aim to obtain a 6 DoF pose by localizing against a map built from multiple images. In general, such maps are built using Structure-from-Motion (SfM) pipelines \cite{Colmap}, which are able to recover the 3D structure from multiple views. Given that this reconstruction is accurate up to a scale factor, additional prior information, such as knowledge of a stereo baseline or direct depth sensing, is required for metric localization. 

For localization, visual place recognition is implemented as an image retrieval problem. Classic methods such as FAB-MAP~\cite{fabmap2008} and DBoW~\cite{DBow2012} solve the problem using local features, but more recent methods such as NetVLAD \cite{Netvlad2018}, PatchNetVLAD \cite{patchnetvlad2021}, EigenPlaces \cite{Berton2023EigenPlaces}, MixVPR~\cite{ali2023mixvpr} rely entirely on neural architectures to obtain global descriptors for retrieval.
Metric pose estimation is then performed via local feature matching using approaches such as SIFT~\cite{lowe2004sift}, SuperPoint~\cite{SuperPoint2018}, or R2D2 \cite{revaud2019r2d2}.
This type of pipeline has been implemented in methods such as HLoc \cite{sarlin2019coarse} and \cite{sattler2017}, which employ a hierarchical approach to large-scale visual localization. 

In addition to the aforementioned methods, alternative approaches have been explored in visual localization in more recent studies. MeshLoc \cite{Panek2022ECCV} shows it is feasible to construct a dense 3D mesh model from multi-view stereo point clouds and to render synthetic images via an OpenGL pipeline to create a visual database. However, a key limitation of MeshLoc is that the retrieval step relies on a database of real images, again limiting the specific location and direction of the localisation.
In another study, \cite{Xue2022CVPR} opts for a different retrieval approach. Rather than using a global descriptor for a query image, they detect instances of buildings in outdoor environments and retrieve the best match from a database of buildings. They show improvement in long-term and large-scale localization datasets over classic hierarchical frameworks.

Works by Trivigno et al.~\cite{MCloc2024} and Zhang et al.~\cite{viewSynthesis2021} employed an iterative strategy to render images and to refine the pose set based on a query image. While this is an effective approach for solving single query images, it requires an initial pose estimate and lacks the real-time capability needed for robotic applications.

\subsection{Lidar Localization}

Similar to visual localization, lidar localization also maintains a map database---in this case of 3D lidar scans. A hierarchical approach to retrieval and matching typically relies on generating global descriptors from each lidar scan. Early work started with handcrafted descriptors. Himstedt et al.~\cite{Himstedt2014GLARE} draw inspiration from the bag-of-words approach and transform 2D lidar scans into a histogram representation for place recognition. He et al.~\cite{He2016M2DP} project the 3D point cloud into multiple 2D planes, generating descriptors for each of the planes based on point density, and combining them into a global descriptor. More recent work has moved towards learned global descriptors. Vidanapathirana et al.~\cite{vid2022logg3d} and Uy et al. ~\cite{uy2018pointnetvlad} both leverage deep learning networks and are able to produce robust global descriptors for loop closure in outdoor large-scale environments.

Recent work in lidar localization techniques has increasingly integrated semantics and extracted segments within the lidar point cloud. Kong et al.~\cite{SemanticGraph2020} introduce a place recognition approach based on a semantic graph, preserving the topological information of the point cloud. They show that by working on a semantic level, their method can be more robust to environmental changes. Aiming to address the same challenge, Segmap~\cite{Dube2020} introduces a segment-based map representation and generates descriptors for these segments using a learned network. These descriptors serve multiple purposes: localization, map reconstruction, and semantic information extraction. Building upon the segment concept, Locus~\cite{Locus2020} further extends the approach by combining descriptors of all segments with spatio-temporal high-order pooling to generate fixed-size global descriptors, which are effective for place recognition. This method was demonstrated to achieve robustness in challenging scenarios such as viewpoint changes and occlusions.
However, there are unique challenges for indoor environments, as objects can clutter the room in confined spaces, leading to more occlusions. InstaLoc~\cite{instaLoc2023} demonstrated good performance for indoor localization by extracting object instances and generating a descriptor for each object before globally matching these objects to estimate the pose.

\subsection{Combined Visual/Lidar Localization}
In scenarios where multiple sensors are available on a mobile robotic platform, some approaches fuse information from both cameras and lidars for localization within a combined image and lidar database.
Oneshot~\cite{Ratz2020OneShotGL} customizes a network to generate descriptors from both lidar segments and their corresponding images from a camera. It estimates the sensor pose by extracting segments from a lidar scan and matching their descriptors to a database. The study demonstrates an enhanced retrieval rate when integrating visual information compared to lidar data alone.
Bernreiter et al.~\cite{bernreiter2021spherical} utilizes a spherical representation to generate descriptors for associated lidar scans and images. A notable advantage of their approach is the flexibility in accommodating different camera and lidar sensor configurations during both training and querying stages.

In recent work, Adafusion~\cite{adafusion2023} introduces a method that employs adaptive weighting on image and lidar pairs to generate descriptors through an attention network. An adaptive weight allows for different contributions from each sensor, and the results show improved retrieval rates and robustness to changing environments.
LC2~\cite{lee2023lc2} is an alternative method wherein 2D images and 3D lidar point clouds are both converted into 2.5D depth images. Then, a network is trained to extract global descriptors from disparity and range images. By reducing modality discrepancy, their method can perform well in very different lighting conditions across multiple missions.

\subsection{Cross-Modal Localization}
For cross-modal localization, most previous work uses a camera to localize in a lidar map. These works are the most relevant to our paper.
EdgeMap~\cite{edgemap2010} extracts straight lines from the point cloud map to build a 3D edge map, and applies an edge filter for the camera image. Based on a particle filter, the likelihood of each pose hypothesis is calculated through the filtered camera image and the edge map.
Building upon the line extraction idea, Yu et al.~\cite{MonoLineCorres2020} extract lines from both the camera image and the lidar map. With a predicted pose from a visual odometry (VO) system, the camera pose is iteratively optimized by minimizing projection error based on 2D-3D line correspondence. They demonstrate that their method can greatly reduce drift by registering live images to the lidar map. 
However, these methods primarily apply to structured environments where lines and edges are prominent features.

Within the context of autonomous driving, researchers have also been developing localization approaches utilizing inexpensive and readily available camera sensors. Wolcott et al.~\cite{visualLoc2014} propose a method to generate a greyscale synthetic view from a mesh map through shading. By minimizing the normalized mutual information between live and synthetic images, this method estimates a pose within the lidar prior map. Similarly, Pascoe et al.~\cite{visualLocCalib2015} synthesize color images from a color textured mesh and can further compensate for camera calibration changes. Synthetic images produced by this (ten year old) work was rather limited, and localization success was constrained to the vehicle's path with fixed viewing angles.

Very recently, more improvements have been made to enhance cross-modal image-to-lidar registration methods. Approaches like Zuo et al.~\cite{StereoLidarMap2020} register dense stereo depth to a lidar map to correct for visual odometry drift. Based on semantic scene understanding, \cite{SemLoc2022} utilizes the point cloud map and its semantic labels to generate a prior semantic map. During runtime, semantic cues are extracted from live images to localize within the prior semantic map.
Both methods fall within the context of camera-lidar registration, which require an initial guess---a relative pose prior. In our work, we focus on global localization without any pose prior information.

\section{Method}

\subsection{Problem Definition}

\newcommand{\tran}{\mathbf{p}}
\newcommand{\World}{\mathtt{W}}
\newcommand{\world}{\mathtt{{W}}}
\newcommand{\camera}{\mathtt{C}}
\newcommand{\map}{\mathtt{M}}
\newcommand{\base}{\mathtt{{B}}}
\newcommand{\T}{\mathbf{T}}
\renewcommand{\S}{\mathtt{S}}

Our goal is to estimate the position and orientation of a single live RGB camera in a prior map, $\mathcal{M}=\{\mathcal{I}_1, \mathcal{D}_1, \mathcal{I}_2, \mathcal{D}_2, ..., \mathcal{I}_n, \mathcal{D}_n\}$, where $\mathcal{I}_i, \mathcal{D}_i$ is the $i$-th synthesized RGB and depth image pair, that are rendered at $\T_{\world\map_i}$. The relevant frames are the earth-fixed world frame $\World$, and the map image frame $\map_i$. 

Unless specified otherwise, the position $\tran_{\world{\map_i}}$ and orientation
$\rotation{\world}{\map_i}$ of the map image are expressed in world coordinates, with 
$\T_{\world\map} \in \SEthree$ as the corresponding homogeneous transform.\noindent

We aim to determine the pose of the live camera image $\mathcal{C}$ at a given time relative to a prior map image, defined as follows:

\begin{equation}
\T_{\camera{\map_i}} 
\triangleq [\mathbf{t}_i, \mathbf{R}_i] 
\in \SOthree
\times \mathbb{R}^{3}
\end{equation}

where $\mathbf{t}_i \in \mathbb{R}^3$ is the translation and $\mathbf{R}_i \in \mathrm{SO(3)}$ is the orientation of $\mathcal{C}$ relative to $\mathcal{M}_i$.
Given $\T_{\world{\map_i}}$ is known, we can find the live camera image pose in the world frame $\T_{\world\mathtt{C}}$.

\subsection{System Overview}

\begin{figure*}
    \centering
    \begin{subfigure}{\linewidth}
        \centering
        \includegraphics[width=0.9\linewidth]{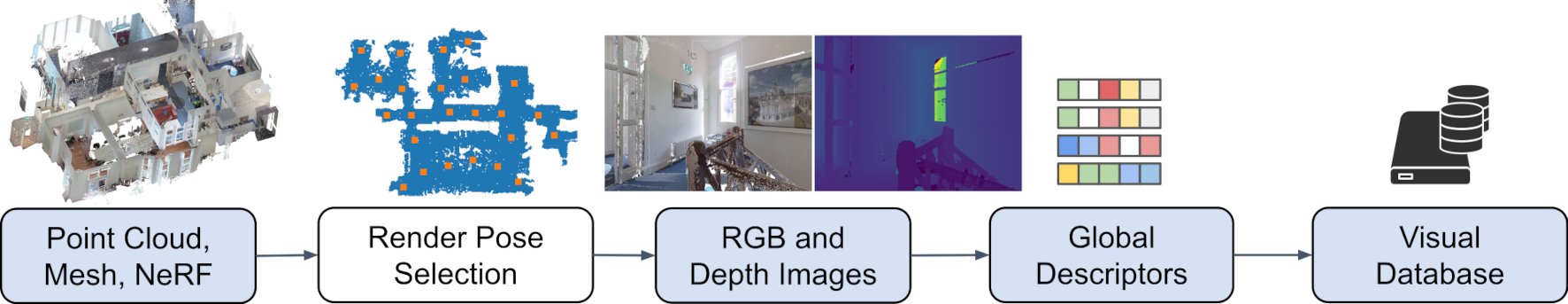}
        \caption{Step 1: Constructing a visual database from a color 3D map.}
        \label{fig:system_a}
    \end{subfigure}\\[15pt]
    \begin{subfigure}{\linewidth}
        \includegraphics[width=0.9\linewidth]{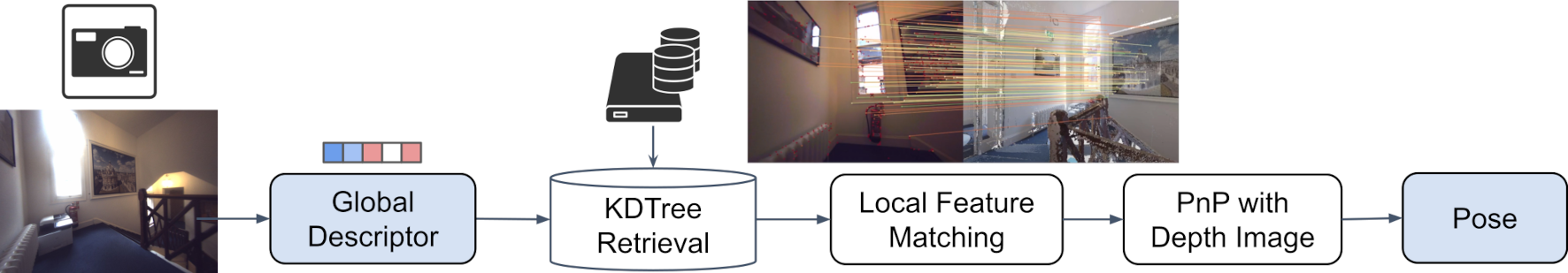}
        \caption{Step 2: A live camera image being matched against a rendered image to obtain a relative pose estimate}
        \label{fig:fig:system_b}
    \end{subfigure}
    
    \caption{Overview of the system showing localization of a camera in the 3D prior map. The blue boxes represent data, and the white boxes represent algorithms.}
    \label{fig:img_loc_overview}
\end{figure*}

We propose a method that integrates learning-based approaches with classical visual geometry to achieve image localization within a color lidar 3D map. The system design is illustrated in \figref{fig:img_loc_overview}, which shows that the system uses a color 3D map (either point cloud, mesh or NeRF based) to generate synthetic RGB and depth images (Step 1). Using these images, we then construct a database of global descriptors. During live operation (Step 2), the system receives a query image and generates a descriptor. It then retrieves the closest match in the database. Learned local features are  extracted from the images and matched. After retrieving the corresponding depth image, the camera pose can then be estimated based on the matched image features and their positions in the world frame.

\subsection{Pose Selection when Rendering a Visual Database}
\label{subsec:pose_selection}
\begin{figure}[]
	\centering
	\includegraphics[width=\linewidth]{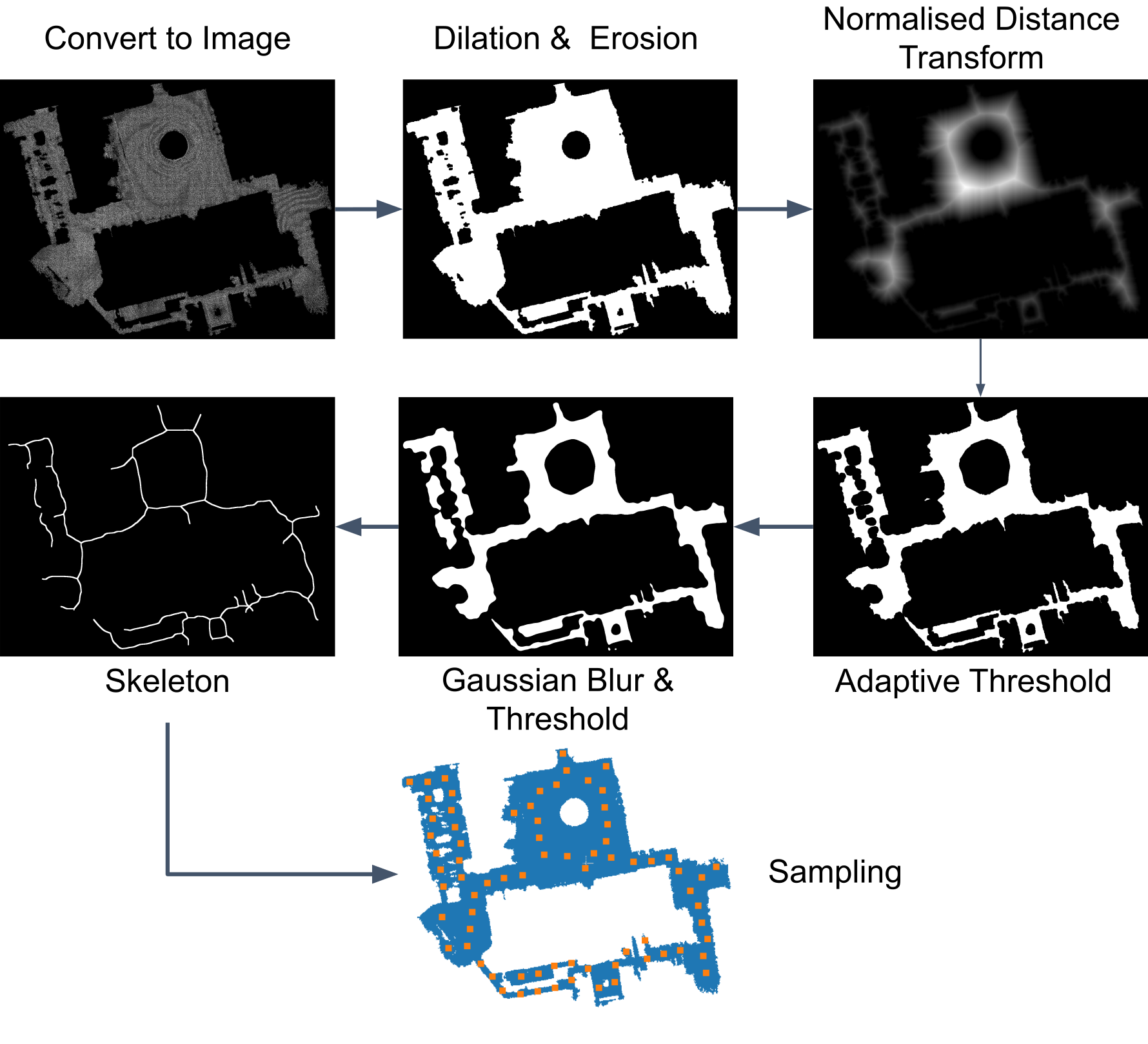}
	\caption{Steps to establish a set of plausible rendering positions within a 3D map, which we call a ``free path corridor''. The map is split into floors and top-down images are rendered containing all the upward-facing points (selected using their normals). The orange points indicate the final selected positions.}
	\label{fig:generate_render_pose}
\end{figure}

A key step is how to decide the set of poses in the 3D prior map from which the artificial images should be synthesized. The goal is to render a reasonably low number of images (to keep the map database small) while achieving coverage across the point cloud or mesh. This step is crucial because it can directly affect both the database size and the available camera views, which influences the overall localization performance. Optimal place recognition can be compromised if the virtual camera is positioned in an unsuitable location---where the map is incomplete, or the camera does not visualize during localization. Our approach involves automatically identifying free space within the map and strategically selecting rendering poses along a ``free path corridor''.


To address this, we propose a straightforward yet effective method that utilizes geometry and image processing to select rendering poses.

For multi-floor building scans, we calculate normals using data from the TLS scans or the SLAM point cloud map. Planes are then extracted by filtering based on normal directions. Each floor plane can be isolated by constructing a histogram counting the number of points with normals facing upwards by point height. Subsequently, we convert each floor plane point cloud into a top-down image and subject this image to a sequence of image processing techniques to determine the reasonable walking area, as shown in \figref{fig:generate_render_pose}. Dilation and erosion operations are applied to close small holes, followed by a normalized distance transform to identify the center of all the free spaces. A smoothed version of the primary ``free path corridor'' is produced using thresholding and Gaussian blurring. Finally, a thinning step is implemented to derive a skeleton representing the path of all the walkable space, which is then sampled to obtain a set of rendering positions. 
We generate four RGB/depth camera image pairs for every position along the path --- forward, back, and two side-facing views. This method is used for all of the 3D prior map representations: point clouds, meshes, and NeRF representations.

\subsection{Generating Synthetic Images}

We aim to generate synthetic RGB and depth image pairs rendered from selected poses using one of our three prior map representations. This section describes the construction of each map representation which in turn facilitates the generation of synthetic images for localization.

\subsubsection{Render Images from Point Clouds}
\label{sec:pc_render}

Point clouds are point samples represented within a 3D coordinate system, sampling an object's or scene's external surface. Point clouds are commonly produced by technologies such as lidar scanning and TLS. In our study, we do not utilize point clouds derived from a SLAM system (such as FastLIO \cite{fastlio2}), we instead use our TLS-generated point clouds as they are also used to generate precise ground truth maps with millimeter-level measurement accuracy. Further details about the TLS scanning process are provided in Sec. \ref{subsec:gt_maps}.

To generate RGB and depth images from the point cloud, we utilize the rendering capabilities of Open3D \cite{Zhou2018}, which leverages an OpenGL framework\footnote{OpenGL, Khronos Group, \url{https://www.opengl.org/}}. Optimal renders are achieved by strategically positioning several light sources around the scene to facilitate specular reflections and shininess. Cameras are placed and maneuvered within the point cloud according to the predefined rendering poses (from the previous subsection). For each pose, 3D points are projected into the camera's field of view using a pinhole projection model, which allows the pixel size of each point to be adjusted.

\begin{figure}
	\centering
	\includegraphics[width=\linewidth]{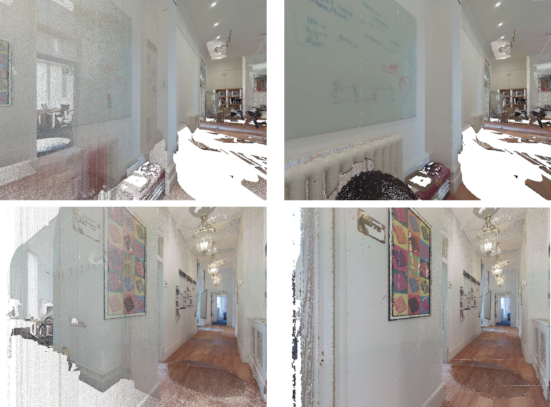}
	\caption{Images from the left and right sides illustrate the before and after of rendered images from the point cloud when utilizing an adjustable point size based on the distance to the virtual camera.}
	\label{fig:adjustable_point_size}
\end{figure}

Positioning the camera close to a wall causes gaps between points when using a small fixed pixel size, leading to a see-through effect in the rendered image, as illustrated on the left side of \figref{fig:adjustable_point_size}. To address this issue, the pixel size of rendered points is varied according to an inverse depth strategy , as detailed in \eqref{eq:point_size_euqation}. This approach ensures that the pixel size remains within the designated maximum and minimum splatting sizes. $\mathbf{z}_{(u,v)}$ represents the depth of a point in 3D space relative to the camera frame.

\newcommand{\pmax}{\mathbf{\rho}_{max}}
\newcommand{\pmin}{\mathbf{\rho}_{min}}
\newcommand{\pixeldepth}{\mathbf{z}_{(u,v)}}

\begin{equation}
    \mathbf{\rho}_{(u,v)} = \min[\max[\frac{\pmax}{\pixeldepth}, \pmin], \pmax]
    \label{eq:point_size_euqation}
\end{equation}

\begin{figure}
	\centering
	\includegraphics[width=\linewidth]{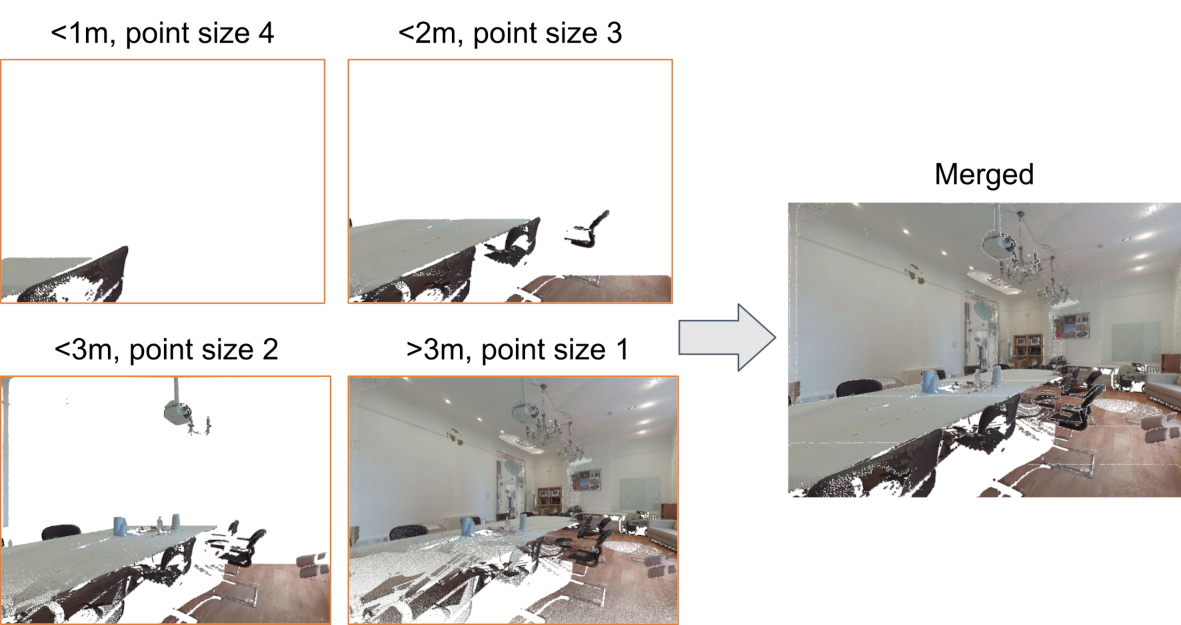}
	\caption{Illustration of merging point cloud rendered images of different point sizes to get the final RGB and depth images.}
	\label{fig:merge_rgb}
\end{figure}

Due to a limitation imposed by OpenGL, adjusting individual point sizes dynamically is not feasible, as the parallel processing architecture requires a uniform point size for effective splatting. To circumvent this limitation, we applied \eqref{eq:point_size_euqation} by rendering multiple frames, each with a point size determined by the maximum and minimum point depths within the scene. These frames are then combined to compose the scene's final RGB and depth image. This is illustrated in \figref{fig:merge_rgb}, showcasing how four images rendered with different point sizes are merged into a single RGB image.

\subsubsection{{Render Images from Textured Meshes}}

Meshes are widely used in robotics as dense representations for 3D scenes and objects. A mesh is a set of point vertices and the edges between them which together form a set of polygonal faces. Textured meshes are a popular method for representing static 3D scenes \cite{arvo2013graphics, botsch2010polygon}, as they allow surfaces to be textured with color images, accurately capturing the scene's appearance.

In our experiments, we extend our previous framework on mesh reconstruction, ImMesh \cite{lin2023immesh}, to reconstruct the triangular mesh of scenes captured using lidar scans. ImMesh initiates this process by estimating the pose of the lidar, followed by registering each scan to a global map. This map is partitioned into fixed-sized volumetric voxels. Mesh reconstruction is carried out online using an incremental, voxel-wise meshing algorithm. Initially, all points within a voxel are projected onto an estimated principal plane to reduce dimensionality. Subsequently, the triangular mesh is reconstructed through a series of voxel-wise operations: mesh pull, commit, and push steps. For more detailed information about the mesh construction process, readers are referred to Section VI of \cite{lin2023immesh}. ImMesh is designed to efficiently reconstruct the mesh of large-scale scenes in real time while maintaining high geometric accuracy.

However, further modifications were necessary to color and texture the mesh reconstruction. While ImMesh constructs a mesh of a scene with adjacent triangle facets connected by edges, these facets lack color information. To address this issue, we use color images captured by the visual camera to texture the facets. 
The camera poses and image exposure time can be estimated using R$^3$LIVE++ \cite{r3live_pp}. To enhance the smoothness and natural appearance of the mesh texture, for each triangle facet $\boldsymbol{\mathcal{T}}_i$, we blend images captured by the $n$ nearest viewing cameras to form its texture $\mathbf{I}_i$ (in our experiment, $n$ is set to 5). The pixel value $\mathbf{I}_i\left( u, v \right)$ at position $(u, v)$ is blended as follows:

\begin{align}
	\mathbf{I}_i \left( u, v \right) = \dfrac{1}{\bar{\sigma}} \sum_{j=1}^{n}{ \Big( \sigma_j \cdot \mathbf{I}_j \left( \boldsymbol{\pi}( \mathbf{R}_j, \mathbf{t}j, \mathbf{P}) \right) \Big) } \\
	\bar{\sigma} = \dfrac{1}{n} \sum_{j=1}^{n}{\sigma_j}, ~~ \mathbf{P} = \mathbf{A}^{-1}( u_i, v_j) \in \mathbb{R}^3
\end{align}
where $\boldsymbol{\pi}(\mathbf{R}_j, \mathbf{t}_j, \mathbf{P})$ is the camera projection function that projects a 3D point $\mathbf{P}$ (on the triangle facet $\boldsymbol{\mathcal{T}}_i$) to the image plane using the estimated camera pose $(\mathbf{R}_j, \mathbf{t}_j)$. $\sigma_j$ denotes the estimated exposure for the $j$-th image, and $\mathbf{A}(\cdot)$ is the wrap transform that maps $\mathbf{P}$ on facet $\boldsymbol{\mathcal{T}}_i$ to $(u, v)$ in texture coordinates, which can be written as $(u, v) = \mathbf{A}(\mathbf{P})$.

After reconstructing the textured mesh, we generate synthetic images of the captured scene. This process is accomplished by rendering the mesh into RGB and depth images using OpenGL with the predetermined rendering poses (the approach described in \ref{subsec:pose_selection}).

\subsubsection{Render Images from Neural Radiance Fields}

NeRF is an emerging 3D representation that is particularly effective for photorealistic novel view synthesis. NeRF uses an implicit representation, usually a Multi-Layer Perceptron (MLP) \cite{mildenhall2020nerf}, to model a radiance field. The scene is represented as a function whose inputs include a 3D location $\textbf{x}=(x,y,z)$ and a 2D viewing direction $(\theta,\phi)$. The output of the function is a color $\textbf{c}=(r,g,b)$ and volume density $\sigma$. 

The optimization of a NeRF is based on regulating the rendered image using the input image as the ground truth. Given the extrinsics and intrinsics of the camera, an image can be represented as a collection of rays $r(t)=o+td$. The expected color of each ray can be computed from the points sampled along the ray using volumetric integration with quadrature approximation
\cite{max1995optical,kajiya1984ray} as $\hat{c}_u = \sum_{i=0}^{N} w_i c_{i}$, where:

\begin{equation} \label{eq:rendering}
w_i =   \exp{\left(-\sum_{j=1}^{i-1} \delta_j \sigma_j\right)} (1-\exp{(-\delta_i\sigma_i)}).
\end{equation}

Early NeRF formulations were very computationally intensive --- with training typically taking days until convergence. Explicit representations such as voxels \cite{fridovich2022plenoxels,mueller2022instant} and 3D Gaussians \cite{kerbl3Dgaussians} have been shown to accelerate training and rendering substantially. Nerfstudio \cite{tancik2023nerfstudio} is a well-supported open-source project which incorporates these representations. It also includes features that are effective when working with real-world data, namely scene contraction~\cite{barron2022mipnerf360} which can better represent unbounded scenes, and appearance encoding~\cite{martinbrualla2020nerfw} which can model per-image appearance changes including lighting conditions and weather. 

As with traditional visual 3D reconstruction systems such as Multi-view Stereo, it is difficult for NeRF to reconstruct textureless surfaces or locations with limited multi-view input. However, accurate depth is important for retrieval and matching, as described in \secref{retrieval_matching}. In addition, training a single NeRF for a large-scale scene is difficult due to 
relative small model size and limited computational hardware.

In this work we use our previous work on visual-lidar NeRF mapping called SiLVR (Scalable Lidar-Visual Reconstruction)~\cite{tao2024silvr}.
SiLVR builds upon the vision-based Nerfstudio pipeline by adding strong geometry regularization using the lidar depth measurements. 
With volumetric rendering, one can render both the expected depth and surface normal. SiLVR uses depth regularization \cite{deng2022depth} to encourage the ray distribution to follow a narrow normal distribution by minimizing the KL-Divergence between them:
\begin{equation}
 \label{eq:depth}
\mathcal{L}_{\text {Depth }}=\sum_{\mathbf{r} \in \mathcal{R}} \mathrm{KL}[\mathcal{N}(\mathbf{D}, \hat{\sigma}) \| w(t)]    
\end{equation}

Additionally, SiLVR adds surface normal regularization to enhance the surface reconstruction. We use the same loss function as MonoSDF~\cite{Yu2022MonoSDF}, but instead of learning-based surface normal prediction for supervision, we estimate surface normals directly using the lidar range image. The rendered surface normal is computed as the negative gradient of the density field and is supervised using the following loss function:

\begin{equation}
	 \label{eq:normal}
    \mathcal{L}_{\text {Normal }}=\sum_{\mathbf{r} \in \mathcal{R}}\|\hat{N}(\mathbf{r})-\bar{N}(\mathbf{r})\|_1+\left\|1-\hat{N}(\mathbf{r})^{\top} \bar{N}(\mathbf{r})\right\|_1
\end{equation}


When training the NeRF model, there are several other considerations. Training large-scale scenes (such as Blenheim Palace in \figref{fig:blenheim_loc}) usually leads to inferior results compared to a smaller scene. One reason for this is the limited model size --- the representation power of our learnt model is determined partially by the number of parameters that can be optimised. A relatively small model trained with a large dataset can lead to under-fitting.  
Additionally, loading thousands of images from a large-scale scene into RAM memory is not always practical. Therefore, SiLVR adopts a submapping system to divide the scene into smaller overlapping submaps. Specifically, the submaps are created by applying Spectral Clustering \cite{von2007tutorial} to divide the full global trajectory into smaller sections. The submaps overlap --- with part of the global trajectory reused at the submap boundaries. This allows a smooth transition from one submap to another.

\subsection{Retrieval and Matching}
\label{retrieval_matching}

Upon completion of the synthetic image generation process and the subsequent
creation of corresponding depth images, the resulting image database is then input into the NetVLAD \cite{Netvlad2018} network. This network is built with a convolutional neural network and a special VLAD layer. The central element of VLAD layer is inspired by the image representation technique known as the Vector of Locally Aggregated Descriptors. The network generates descriptors capable of accurately identifying the location of the query image, even amidst significant clutter (e.g., people, cars), variations in viewpoint, and stark differences in illumination, including day and night conditions. The descriptors of all the rendered RGB images are stored in the KDTree data structure for future indexing.

Upon receiving an incoming live camera image, our system initially undergoes a distortion correction process before being subjected to the same NetVLAD network. This step produces a descriptor for the live camera image. Afterward, the KD-Tree retrieves the virtual image that exhibits the closest match based on the descriptor values.

For the pair of matched cameras and virtual images obtained in the aforementioned step, local features are extracted using the SuperPoint algorithm. Due to the large domain gap between the synthetic-to-real images, we observe traditional, not learning-based feature detectors, such as SIFT \cite{lowe2004sift}, fail to detect common features. 
SuperPoint is first pre-trained on a synthetic dataset, boosting the domain adaptation performance. Then SuperPoint is further trained with a multi-scale, multi-transform augmentation, enabling self-supervised training of interest point detectors. This gives SuperPoint the properties of repeatable feature detection across different view angles. 

Subsequently, these local features are matched through SuperGlue \cite{SuperGlue2020}. The SuperGlue network constitutes a Graph Neural Network coupled with an Optimal Matching layer, which is trained specifically to conduct matching on two sets of sparse image features. 

Finally, with access to the matched local features and the depth image, the Perspective-n-Point (PnP)\cite{PnPSurvey} algorithm is used to estimate the relative pose between the live camera image and the rendered reference image. Based on the known pose of the reference image, we can triangulate the live query image in the color 3D map. The pseudo algorithm describing the details of the matching process is shown in Algo. \ref{algo:pose_estimation}; note that we utilize the confidence score from the SuperGlue network to filter out bad matches.

\renewcommand{\algorithmicrequire}{\textbf{Input:}}
\renewcommand{\algorithmicensure}{\textbf{Output:}}
\begin{algorithm}
	\caption{Query pose estimation with reference image.}
        \label{algo:pose_estimation}
	\begin{algorithmic}[1]
            \Require{camera intrinsic $f_x, f_y, c_x, c_y$, depth image $D$, reference image pose in world frame $T_{WR}$} 
            \Ensure{$T_{WQ}$ (query image camera pose in world frame)}
            \For {$i=1,2,\ldots,N$ matched keypoint}
                \State $q_u, q_v \gets$ query image keypoint[i]
                \State $r_u, r_v \gets$ reference image keypoint[i]
                    \State depth $P_z \gets D[v][u]$
                    \State $P_x \gets (r_u - cx) * z / fx$
                    \State $P_y \gets (r_v - cy) * z / fy$
                    \If {match confidence score $\omega >0$}
                    \State 3d points $\gets[P_x, P_y, P_z]$
                    \State image points $\gets[q_u, q_v]$
                \EndIf
            \EndFor 

            \If { image points $\leq$ 6}
                \State $T_{QR}$ , success $\gets$ perspective N points with (3D points, image points) 
                \If { success }
                    \State Return $T_{WQ} =  T_{WR} * T_{QR}^{-1}$
                \EndIf
            \EndIf
	\end{algorithmic}
\end{algorithm}

\subsection{Implementation Details}

For NeRF 3D reconstruction, we train each SiLVR map over 150,000 iterations, with 4096 rays sampled from the training set in each iteration. This takes around 5 hours on average for all experiments on a single Nvidia RTX 4090 GPU.

For the localization system, the input RGB images have 720$\times$540 pixels resolution, and we use off-the-shelf pre-trained networks. Specifically, we use a pre-trained NetVLAD backbone based on VGG16, trained on the Pitts30K \cite{Torii_2015_pitts} dataset. Additionally, the SuperGlue network we use is pre-trained on ScanNet data \cite{dai2017scannet} for indoor scenes and MegaDepth data \cite{MegaDepthLi18} for outdoor environments.

The localization part of the system (Step 2) is designed for real-time applications. Upon initialization, all rendered images are loaded, global descriptors are extracted for KDTree indexing, and local features are stored in memory. All algorithms are implemented on a Dell laptop with an Intel Core i7-9850H CPU running at 2.60GHz and a Nvidia Quadro T2000 GPU with 4GB of memory. The online localization pipeline processes each input image in approximately 0.5 seconds, with the following upper limits for each step: global descriptor generation (200ms), KDTree retrieval (1ms), local feature detection (60ms), local feature matching (100ms), and pose estimation (60ms). These performance metrics enable the system to comfortably operate at \SI{1}{Hz}.

\section{Experiment Setup and Hardware}

\subsection{Dataset}

We collected two recordings at each location: one was used to build the prior map and to render the database images, and the second was used for localization testing. The recordings were collected with a walking speed of around 1m/s. The test locations had the following distinct characteristics:
\begin{itemize}
    \item ORI [indoor] was collected at the Oxford Robotics Institute. The ORI dataset encompasses indoor environments such as office rooms, staircases, and a kitchen. This dataset is characterized by faster camera rotation movements than the outdoor dataset with the content in the images changing quickly as the recording device moves from room to room.
    \item Math [outdoor, medium scale] was collected outside the Oxford Mathematics Institute on a summer's day. This dataset is characterized by substantial lighting variations, with transitions between areas of direct sunlight and shadows cast by the buildings. The presence of bushes, trees, and lawns increases the complexity of the image rendering.
    \item Blenheim [outdoor, large scale] was collected at Blenheim Palace, a 300-year-old country house in Woodstock, Oxfordshire. It captures walking sequences within the palace courtyard, partially extending into the palace's main hall. This dataset is challenging as images often contain significant portions of ground and sky, while the geometric symmetry of the courtyard further adds complexity.
\end{itemize}
The test recording, \texttt{ORI}, was collected across 7 office rooms spread over 2 floors with a total trajectory length of \SI{125}{\meter}. The \texttt{Math} trajectory is \SI{451}{\meter} long, and the \texttt{Blenheim} trajectory is 386m long.

Again please note that the datasets and ground truth maps used here are part of the Oxford Spires Dataset mentioned in Sec. \ref{sec:intro} .
\subsection{Hardware - Frontier}

\begin{figure}
    \centering
    \includegraphics[width=\linewidth]{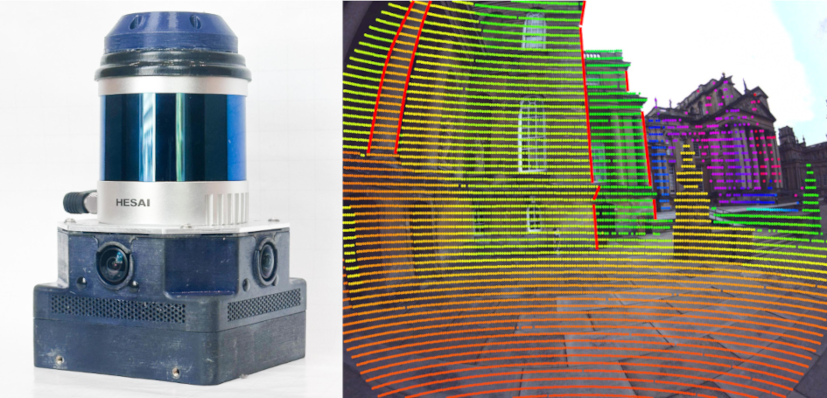}
    \caption{Frontier multi-sensor unit consists of a 3D lidar, 3 orthogonal cameras and an IMU (left image). A fisheye image overlaid with lidar points (right image), demonstrating accurate intrinsic and extrinsic calibration (red lines are manually added to highlight the depth change).}
    \label{fig:hardware}
\end{figure}

All experimental data was collected using a \textit{Frontier}, \figref{fig:hardware}, a self-developed multi-sensor unit consisting of a Sevensense Alphasense Multi-Camera kit with 3 orthogonal cameras of 1440$\times$1080 pixels resolution. The device contains a Hesai QT lidar with 64 beams and \SI{104.2}{\degree} vertical field of view. Note that the Hesai lidar is used to build the mesh and NeRF maps and generate ground truth poses.

The extrinsic calibration between the lidar and camera is established following the methodology outlined in \cite{DiffCal}, ensuring sub-millimeter accuracy in translation and sub-degree precision in rotations. Accurate calibration is crucial, as shown in Fig.~\ref{fig:hardware}, particularly for accurately overlaying lidar scans onto color images when reconstructing color 3D maps.

\subsection{Ground Truth Maps and Trajectories}
\label{subsec:gt_maps}
To generate ground truth point cloud maps, we used a (professional grade) Leica RTC360 Terrestrial Laser Scanner (TLS) to scan the \texttt{Math} and \texttt{Blenheim} sites and an (entry-level) Leica BLK360\footnote{\url{https://leica-geosystems.com/products/laser-scanners}} to scan the \texttt{ORI} site. Views of these maps are shown in the left section of Tab~\ref{tab:full_sys_eval}. TLS scans are captured around \SI{3}{\meter} apart for indoor \texttt{ORI} and \SI{15}{\meter} to \SI{20}{\meter} apart for outdoor \texttt{Math} and \texttt{Blenheim}. Note that ultra-dense distance between TLS scans is impractical due to the considerable time required to scan and process point clouds.

Each Hesai lidar scan was registered to the ground truth point cloud to obtain a 6 DoF pose. We refer readers to our previous paper, where we explain the details of the ground truth generation process in more detail \cite{zhang2021multicamera}, which we claim achieves centimeter precision.

These point clouds were also used to render the synthetic images for the point cloud map representation, as explained in Sec \ref{sec:pc_render}. Before rendering, we voxel-filtered the \texttt{ORI} point cloud to \SI{5}{\milli\meter} resolution and the \texttt{Math} and \texttt{Blenheim} clouds to \SI{1}{\centi\meter}.

\section{Experiment Results}
\subsection{Comparison of Rendered Images}

\begin{figure*}
    \centering
    \includegraphics[width=\linewidth]{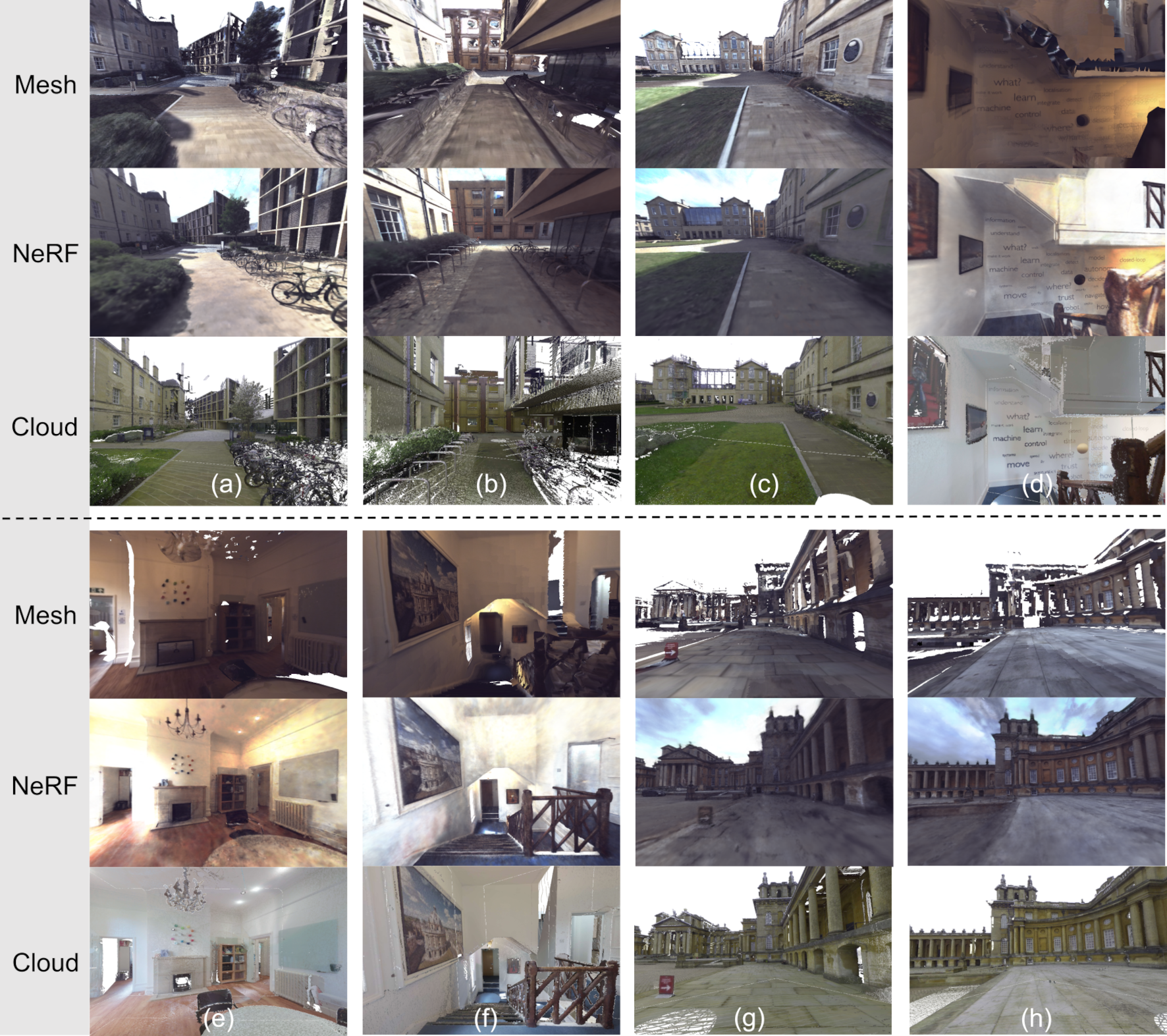}
    \caption{A comparison of rendered images from mesh, NeRF, and point cloud in \texttt{Math} (a-c), \texttt{ORI} (d-f), and \texttt{Blenheim} (g-h).}
    \label{fig:render_image_comparison}
\end{figure*}

First we present a set of illustrative examples of rendered images obtained using the TLS point cloud, NeRF, and mesh reconstruction pipelines. Fig.~\ref{fig:render_image_comparison} illustrates examples drawn from the three datasets. Each pipeline has the ability to generate reasonably lifelike synthetic images. However, we note that each also exhibits distinct limitations.

For the indoor \texttt{ORI} dataset, object occlusion and the relatively sparse coverage of the TLS scanner results in some visible gaps in the point cloud, leading to white regions in the rendered images. Mesh reconstruction faces challenges in accurately representing smaller objects or intricate details, primarily due to the fixed size of the triangular faces. This method is more suitable for objects with planar surfaces than objects with complex profiles. For example, the handrails are difficult to reconstruct with a mesh, as shown in (f).
Conversely, NeRF demonstrates its superiority in outdoor environments, successfully reconstructing fine details such as bicycle rails where mesh reconstruction struggles, as shown in (b). However, NeRF often fails to capture intricate ground floor patterns, as shown in (g) and (h), while mesh reconstruction and point cloud techniques, which employ direct image projection, can easily accomplish this.

\subsection{Full System Evaluation}

\begin{table*}
\centering
\caption{Performance comparison of the different localization methods. ("DB size'' is the number of images in the database. "Ret. Rate'' refers to the retrieval rate. "Loc. Rate'' means the localization rate. COLMAP (*) required all the images from all three cameras to reconstruct the prior map accurately for the indoor ORI recording.)}
\label{tab:full_sys_eval}

\resizebox{\linewidth}{!}{%

\begin{tabular}{c|clcr|rr}
\toprule
\multirow{19}{*}{
\begin{minipage}{0.7\textwidth}
    \centering
    \includegraphics[width=\textwidth]{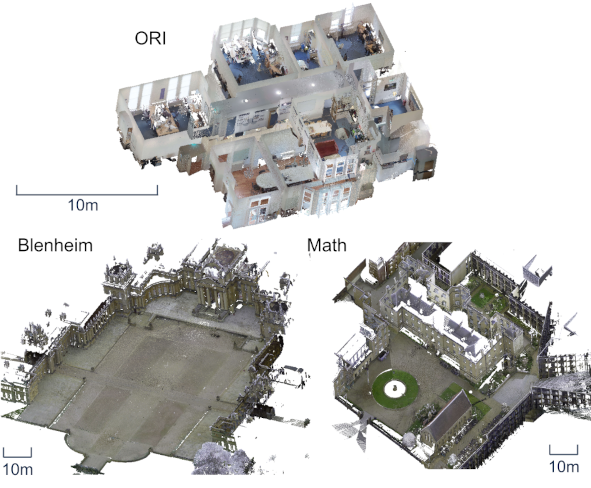}
\end{minipage}
}
                       &\textbf{Dataset}            & \multicolumn{2}{c}{\textbf{Method}}  & \textbf{DB Size}      & \textbf{Ret. Rate} & \textbf{Loc. Rate}     \\ \cmidrule{2-7}
                       &                            &                        & Sparse      & 130                   & N/A                & Fail                   \\ 
                       &                            & \multirow{-2}{*}{HLoc} & Dense       & 254                   & N/A                & 86                     \\ \cmidrule{4-5}
                       &                            & COLMAP                 & Dense       & 3020*                 & 97                 & \textbf{96*}           \\ \cmidrule{3-7}
                       &                            &                        & Cloud       &                       & 68                 & 54                     \\ 
                       &                            &                        & Mesh        &                       & 81                 & 61                     \\ 
                       & \multirow{-5}{*}{ORI}      & \multirow{-3}{*}{Ours} & NeRF        & \multirow{-3}{*}{130} & 86                 & \textbf{67}            \\ \cmidrule{2-7}
                       &                            &                        & Sparse      & 503                   & N/A                & 20                     \\ 
                       &                            & \multirow{-2}{*}{HLoc} & Dense       & 1019                  & N/A                & 39                     \\ \cmidrule{4-5}
                       &                            & COLMAP                 & Dense       & 1569                  & 66                 & \textbf{61}            \\ \cmidrule{3-7}
                       &                            &                        & Cloud       &                       & 89                 & 56                     \\ 
                       &                            &                        & Mesh        &                       & 85                 & 57                     \\ 
                       & \multirow{-5}{*}{Math}     & \multirow{-3}{*}{Ours} & NeRF        & \multirow{-3}{*}{300} & 97                 & \textbf{75}            \\ \cmidrule{2-7}
                       &                            &                        & Sparse      & 210                   & N/A                & 42                     \\ 
                       &                            & \multirow{-2}{*}{HLoc} & Dense       & 449                   & N/A                & 55                     \\ \cmidrule{4-5}
                       &                            & COLMAP                 & Dense       & 449                   & N/A                & \textbf{78}            \\ \cmidrule{3-7}
                       &                            &                        & Cloud       &                       &77                  & 57                     \\ 
                       &                            &                        & Mesh        &                       &77                  & 56                     \\ 
                       & \multirow{-5}{*}{Blenheim} & \multirow{-3}{*}{Ours} & NeRF        & \multirow{-3}{*}{210} &86                  & \textbf{73}            \\ 
\bottomrule
\end{tabular}
}
\end{table*}

\begin{table}
\centering
\caption{Comparative analysis of localization methods for use in different robotics tasks.
{\color{myGreen}\cmark} means suitable, {\color{myOrange}\halfcmark} means partly suitable, and {\color{myRed}\xmark} means unsuitable.}
\label{tab:qualitative_comparison}
\begin{tabular}{lccc}
            \toprule
            \textbf{Method}   & \textbf{Metric Pose}           & \textbf{Real Time}          & \textbf{Planning} \\ \midrule
            HLoc                   &                            &                               &                                \\ 
            COLMAP                 &  \multirow{-2}{*}{\color{myOrange}\halfcmark}   &  \multirow{-2}{*}{\color{myRed}\xmark}      & \multirow{-2}{*}{\color{myOrange}\halfcmark}      \\ \cmidrule{2-4}
            Ours-Cloud             &                              &                                &                               \\
            Ours-NeRF              &                               &                                &                               \\
            Ours-Mesh              &   \multirow{-3}{*}{\color{myGreen}\cmark}     & \multirow{-3}{*}{\color{myGreen}\cmark}       & \multirow{-3}{*}{\color{myGreen}\cmark}      \\ 
            \bottomrule
\end{tabular}
\end{table}

In this section, we quantify the localization performance of the three different rendering pipelines using these synthesized images, benchmarking performance against two entirely vision-based localization systems (which use real images), as detailed in Tab. \ref{tab:full_sys_eval}. The systems we compare are HLoc and COLMAP, both based on Structure-from-Motion (SfM). We selected these two methods for comparison because they are widely adopted in the community and are designed for large-scale real-world reconstructions. Once the SfM step is completed, these methods allow for real-time localization of query images.

Tab.~\ref{tab:qualitative_comparison} presents a high-level analysis of the suitability of these systems for key robotics tasks. The SfM approaches are non-metric, meaning their pose estimates are not scaled to real-world dimensions. This limitation necessitates other information to be available to allow effective deployment in robotic applications. These methods are also not real-time, as the SfM bundle adjustment process would take a couple of hours to a day, depending on the image or map size. 

In the following section, we will discuss the configurations of HLoc and COLMAP in our experiments and compare them with our methods of localizing using point cloud, mesh, and NeRF-based map representations. For a fair comparison, the primary performance metrics we use are \textit{retrieval rate}, the proportion of images correctly retrieved as a percentage of the total number of queried images, and \textit{localization rate}, the proportion of images correctly retrieved and localized within a threshold as a percentage of the total number of queried images. As we are evaluating global localization systems, the threshold is within \SI{1}{\meter} and \SI{30}{\degree} of ground truth poses for indoors \texttt{ORI}. For outdoor \texttt{Blenheim} and \texttt{Math}, the threshold is within \SI{2}{\meter} and \SI{30}{\degree}.

\subsubsection{Our Methods -- Cloud, Mesh and NeRF}

\begin{figure}
	\centering
	\includegraphics[width=\linewidth]{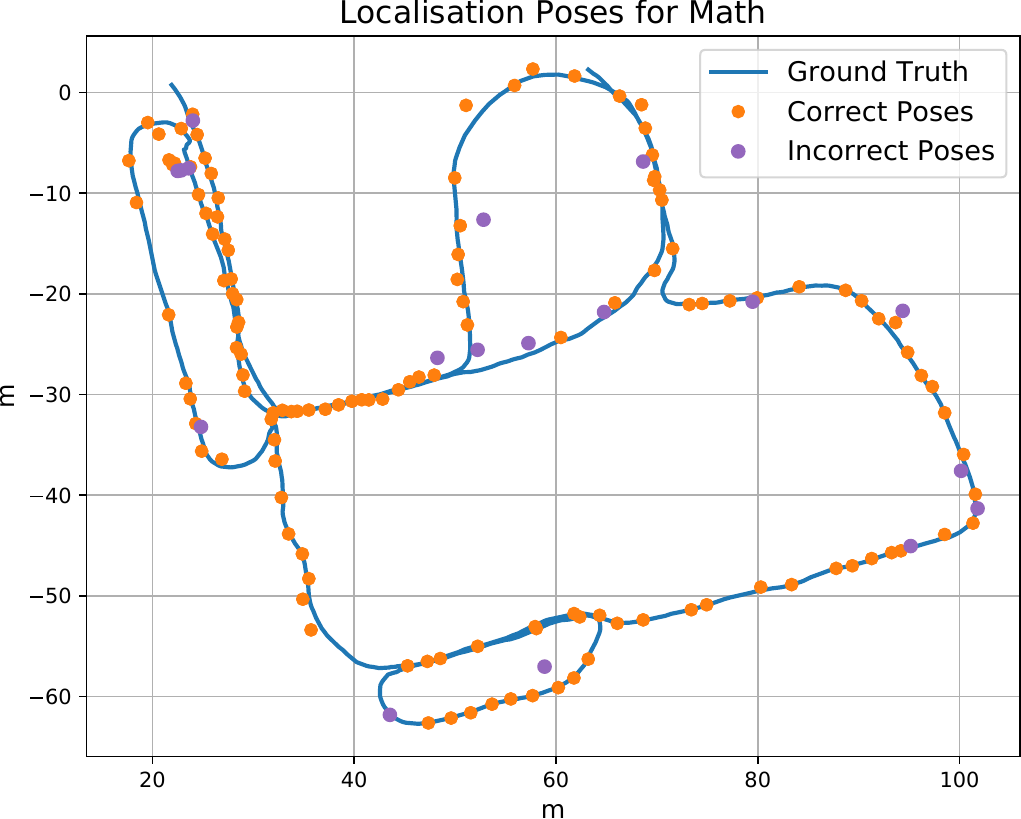}
	\caption{Estimated poses in \texttt{Math} from the NeRF synthesized image database. The purple dots show estimated poses exceeding the \SI{2}{\meter} and \SI{30}{\degree} threshold compared to the ground truth poses.}
	\label{fig:math-loc-result}
\end{figure}

For a fair comparison, we render an equal number of images for each dataset at the same locations. For \texttt{ORI}, render poses are \SI{2}{\meter} apart, generating 130 images. Since \texttt{Math} and \texttt{Blenheim} cover larger areas than \texttt{ORI}, render poses are \SI{4}{\meter} apart, and 300 and 210 images were produced, respectively.

When testing our approach of localizing with a database of rendered images, we found consistent performances for each of our three reconstruction techniques (point cloud, mesh, and NeRF) across three distinct test locations. On average, the localization rates for the point cloud and mesh approaches were \SI{56}{\percent} and \SI{58}{\percent}, respectively, while the NeRF-based approach achieved the best performance at \SI{72}{\percent}. The relatively lower localization rates with point cloud and mesh systems can be attributed to the projection of color and textures from multiple camera views onto the 3D space, occasionally resulting in occluded regions and color inconsistencies.

\figref{fig:math-loc-result} illustrates the performance of pose estimation for query images localized in \texttt{Math} against a NeRF-rendered image database. The ground truth pose trajectory is drawn in blue, the estimated poses in orange dots, and the estimated poses that exceeded the threshold are highlighted in purple.

The notable advantage of point cloud and mesh mapping systems is their ability to generate color 3D maps online. The point cloud and mesh results are available immediately following a SLAM run, making them useful for real-time applications that cannot support rendering delays. Conversely, NeRF reconstruction offers the most photorealistic rendering outcomes but requires training after data collection. Additionally, synthetic image generation with the NeRF reconstruction is executed through an MLP inference step, which also requires a slightly longer time (e.g. $\sim\SI{1}{\second}$ per image of resolution 720$\times$540 on a Nvidia RTX 4090 GPU). In contrast, the OpenGL-based rendering utilized in the point cloud and mesh method takes a few milliseconds per image.

\subsubsection{HLoc}

HLoc \cite{sarlin2019coarse} is a 6 DoF visual localization toolkit. It employs a hierarchical localization approach that integrates image retrieval and feature-matching techniques. We first configured HLoc to build an SfM model using all real images and then used it to register query images to the SfM model. For an objective comparison, we used the same pre-trained networks as used in our localization method --- NetVLAD for image retrieval, SuperPoint for local feature detection, and SuperGlue for feature matching. 

For HLoc and COLMAP, the experiments started with the same number of database images as our system. We increased the number of images until their performance improvement plateaued. In general, visual localization systems require a larger number of images to construct an accurate SFM.

To obtain query image pose estimates (up to the scale metric), we first refine the pose estimates by eliminating any erroneous matches and then use Evo\footnote{\url{https://github.com/MichaelGrupp/evo}} to align the query image poses with the ground truth poses to recover the scaling factor and the alignment transformation. Finally, we apply this scaling factor to all estimated poses. We can then calculate the localization rate. 

During the initial SfM reconstruction step, HLoc performed well with indoor data but encountered difficulties in accurately constructing the larger outdoor environments of \texttt{Math} and \texttt{Blenheim}. We found that constructing a robust SfM solution requires a dense set of images with substantial overlap and minimal lighting changes.
For \texttt{ORI}, an image database of 130 images (as used for map rendering pipelines) was insufficient as the images were too sparse for the SfM solver to achieve sufficient overlap. Increasing the number of images (to 254 and later 500) enabled the creation of a complete reconstruction, albeit with some observable drift in the staircases connecting the two floors of the building.  For \texttt{MATH}, the best results were achieved with 1019 images, and increasing the database (to 2040 images) did not noticeably improve the SfM model. Lastly, for \texttt{Blenheim}, HLoc achieved its best performance with 449 images, and doubling the image count did not improve the overall SFM accuracy.

Overall, HLoc worked well in the indoor setting, but the reconstruction was less successful in large outdoor scenes.


\subsubsection{COLMAP}

This system \cite{Colmap} is a widely-used SfM and Multi-View Stereo pipeline with graphical and command-line interfaces. It can construct visual maps using ordered and unordered image collections. In our experiments, we adopted the recommended approach of utilizing SIFT as the feature detector and a vocabulary tree for image retrieval. Our results show that COLMAP requires a higher number of database images than HLoc. It achieves its best performance when incorporating images from the three cameras of the Frontier (front, left, and right facing) within the dataset.
This was particularly the case for the indoor \texttt{ORI} dataset, where the camera motion is more abrupt/jerky. For \texttt{ORI}, COLMAP required all images from all three cameras --- facing left, forward, and right, a total of 3020 images.

Once the SfM solution is accurately estimated, localizing new images is straightforward, especially if a query image closely resembles the database images. For \texttt{MATH}, the query and map images were captured in quite different lighting conditions. Hence, the performance of image retrieval is lower with COLMAP as it uses the non-learning-based image retrieval method. 
Across all three locations, COLMAP gave good localization performance, but constructing the SFM solution takes double the amount of time compared to HLoc, e.g., for \texttt{MATH}, COLMAP takes around 4.5 hours with 1569 images, and HLoc takes 2 hours for 1019 images.

\subsection{Image Descriptor, Feature Detector and Matcher}
For a fair comparison, we adopted the same image descriptor and feature detector methods as HLoc. We also experimented with image descriptors such as EigenPlaces~\cite{Berton2023EigenPlaces} and MixVPR~\cite{ali2023mixvpr}, but NetVLAD marginally outperformed them when tested on real query images and synthetic database images. A more detailed ablation study comparing learning-based and non-learning-based feature detectors and matchers is provided in Sec. \ref{ablation_study2}. It is important to note that our approach is not dependent on the specific learning-based detector and matcher --- which can be swapped for improved methods as they emerge. The key point here is that by sampling a database of synthetic images, we can leverage dense 3D maps to provide a \textit{unified representation} across different map types. This approach offers flexibility in generating poses within the 3D map and even allows for an iterative "render and compare" strategy if needed, while also reducing the number of database images requires, as compared to SfM-based approaches.

\subsection{Localization in Directions Unseen during Mapping}

\begin{figure}
    \centering
    \begin{subfigure}{\linewidth}
        \includegraphics[width=\linewidth]{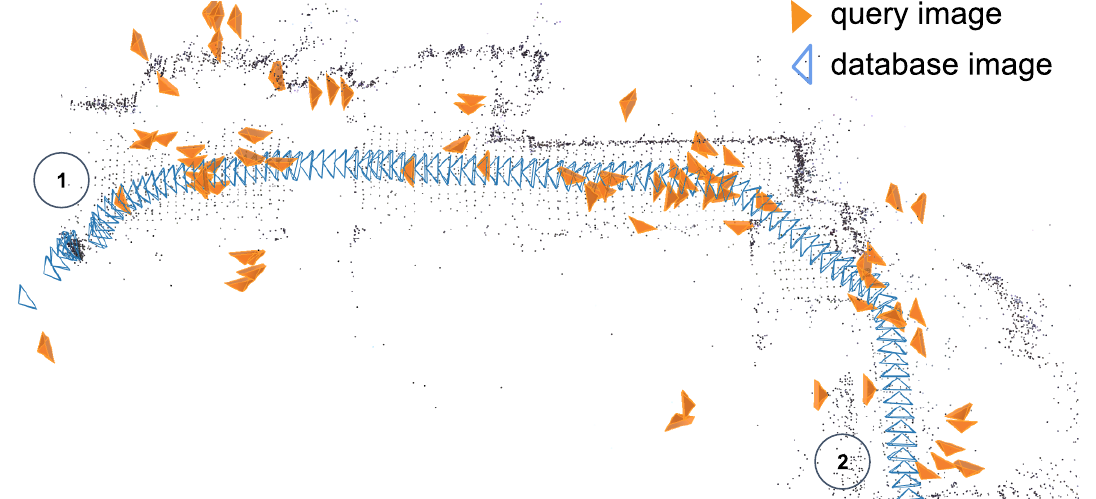}
        \caption{HLoc map: Although the SfM reconstruction is accurate, only four camera views could be correctly localized around the bend.}
        \label{fig:oppo_direction_loc_hloc}
    \end{subfigure}
    \\[10pt]
    \begin{subfigure}{\linewidth}
        \centering
        \includegraphics[width=\linewidth]{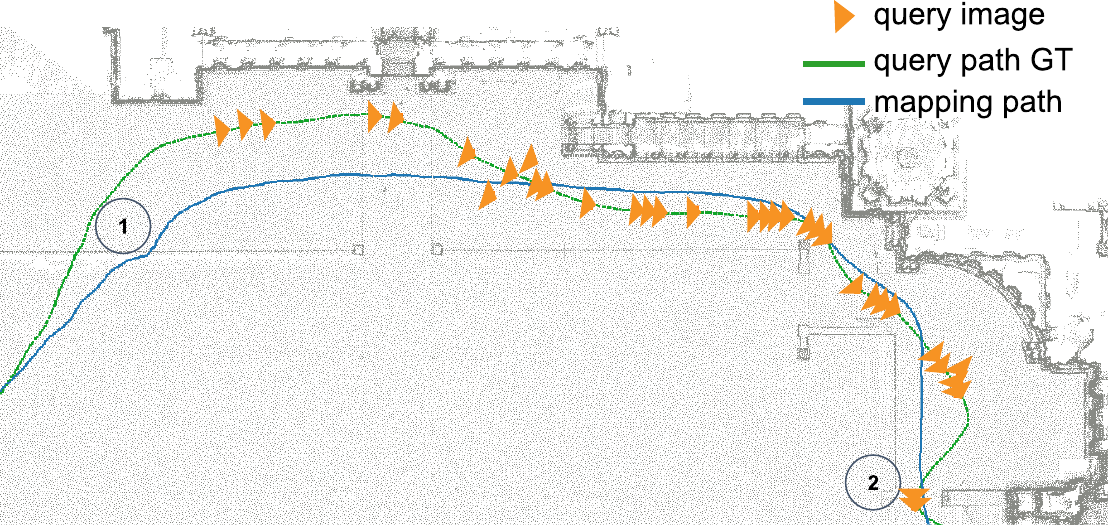}
        \caption{Mesh map: 30 out of 90 query images could be correctly localized.}
    \end{subfigure}
    
    \caption{Localizing when traveling in a direction opposite to the mapping phase. The mapping direction (blue) is from 1 to 2, and the query direction (green) is from 2 to 1. Orange triangles represent the localized camera view results.}
    \label{fig:oppo_direction_loc}
\end{figure}

\begin{table*}
    \centering
    \caption{Localization when traveling in the opposite direction to the mapping trajectory with novel view synthesis.}
    \label{tab:opposite_loc}
    \begin{tabular}{lcrrr}
    \toprule
    \textbf{Method} & \textbf{Camera} & \textbf{Images}   & \textbf{Ret. Rate \%} & \textbf{Loc. Rate \%} \\
    \midrule
    COLMAP & front & 98 & N/A & 0 \\
    HLoc & front   & 98 & N/A & 4  \\
    NeRF & front & 30 & 10 & 7\\
    Mesh & front & 30 & 37 & \textbf{33} \\
    \midrule
    COLMAP & front + side & 196 & N/A & 26  \\
    HLoc & front + side & 196 & N/A & 39 \\
    NeRF & front + side & 30 & 88 & \textbf{75} \\
    Mesh & front + side & - & - & - \\
    \midrule
    Point Cloud & omni & 30 &  98 &  \textbf{88}\\
    \bottomrule
    \end{tabular}
\end{table*}

Our method has the powerful capability of generating images from any viewpoint. This is very useful when attempting to localize when traveling in directions opposite to that used to create the map. For instance, in the scenario where the map is constructed by traversing from left to right, as shown in \figref{fig:oppo_direction_loc}, we can synthesize images from all four orthogonal directions. This capability facilitates localization even when traveling in the reverse direction, a challenge often encountered in conventional visual localization methodologies, which typically requires storing images from both travel directions in the database.

The initial recording was used to reconstruct a mapping database with only the front-facing camera, as shown in \figref{fig:oppo_direction_loc_hloc}. The map was generated by traversing from point 1 to point 2, where 98 images were captured (illustrated in blue). We then attempted to localize to the map during a reverse traversal from points 2 to 1, with the results depicted using orange-colored camera views. Of the 87 query images, only 4 were successfully localized (when turning around the corner) due to minimal overlap in the image views. A similar pattern was observed with COLMAP where it failed to localize any query images even though the localization module had an accurate SfM reconstruction. HLoc performed marginally better than COLMAP due to its learning-based feature matching approach.

NeRF generated blurrier images from reverse perspectives when trained exclusively with front-facing camera images, resulting in suboptimal localization performance. This issue is likely caused by insufficient training views when the path involves walking in a straight line. In contrast, the mesh-based pipeline demonstrated strong performance in this scenario. Mesh could effectively generate rendered images facing in the un-mapped direction (albeit with imperfections resulting in holes in the rendered images). Despite these visual limitations, mesh-rendered images achieved a localization rate of approximately \SI{33}{\percent}.

In a second experiment, we expanded the mapping database to include images from two cameras, the side and front camera. As expected, this enhanced the performance for COLMAP and HLoc, with side camera views available for reverse image matching. NeRF exhibited much better localization performance, matching 75\% of the images. Note that the mesh reconstruction implementation only works with one camera. 

To provide a broader comparison, we show that point clouds captured with an omnidirectional camera (both front and backward facing) can achieve a very high localization rate.

To conclude, this section demonstrates that our system can generate synthetic images in novel views, which aids the localization in a direction that is unseen during the mapping. In particular, mesh-based map representation can render images for localization even when there are very limited views of the environment, while NeRF can render better images for localization with a moderate number of views.

\subsection{Ablation Study: Changes in Environment}

\begin{figure*}
    \centering
    \begin{subfigure}{\linewidth}
        \centering
        \includegraphics[width=\linewidth]{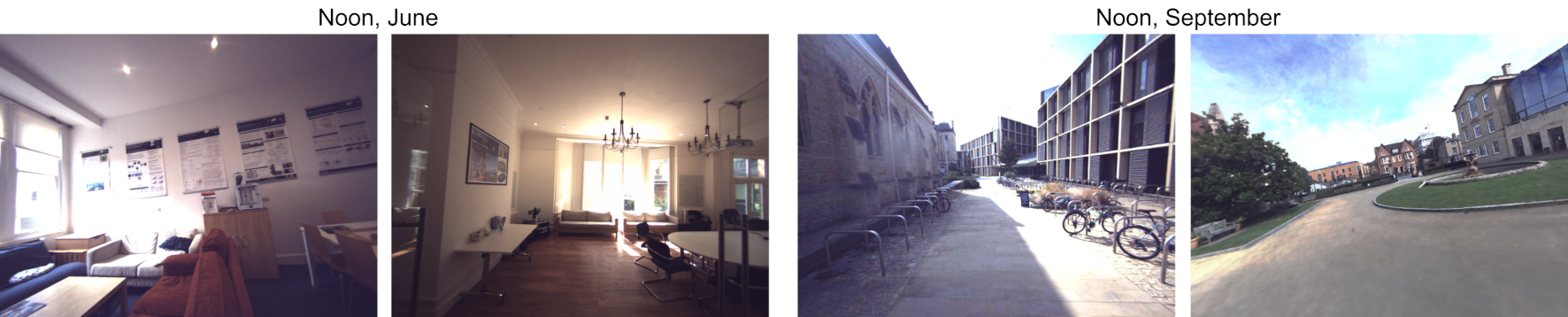}
        \caption{Scenes in the original maps.}
    \end{subfigure}
    
    \begin{subfigure}{\linewidth}
        \includegraphics[width=\linewidth]{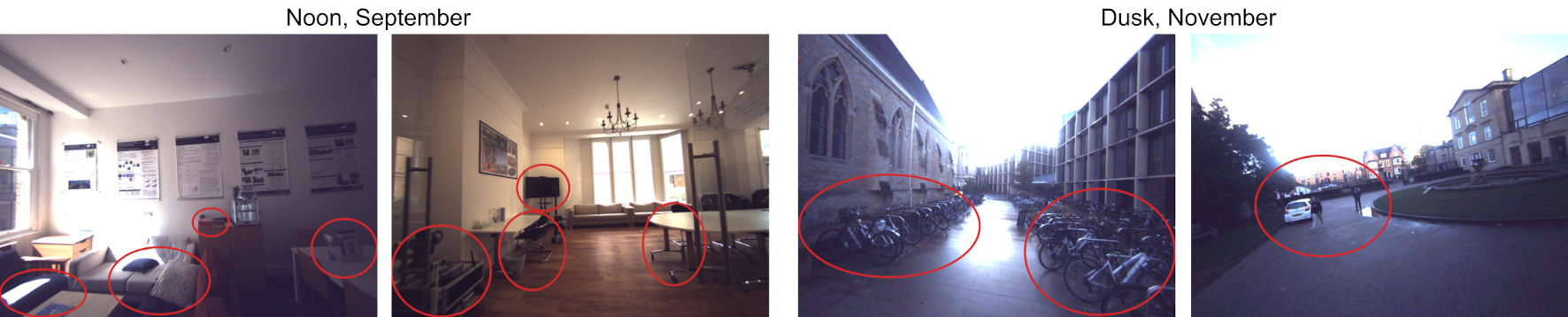}
        \caption{Changes after three months.}
    \end{subfigure}
    
    \caption{Ablation study: assessing changes in \texttt{Math} and \texttt{ORI} 2-3 months apart. The top row displays images from the original dataset, while the bottom row shows corresponding images captured months later which reflect the typical movement of objects within the scene. The images on the right-half are from \texttt{Math}, while the bottom right ones were captured on a dark day. Please note that some images have been brightened for easier interpretation here.}
    \label{fig:changes-in-scene}
\end{figure*}

\begin{table*}[]
\centering
\caption{Localization performance amid scene changes in prior maps.} 
\label{tab:change_in_env}
        \begin{tabular}{ccccc}
        
        \toprule
        \textbf{Image Source}            & \textbf{Dataset}      & \textbf{Time}  & \textbf{Ret. Rate}      & \textbf{Loc. Rate}        \\ \midrule
        \multirow{2}{*}{Point Cloud}     & \multirow{2}{*}{ORI}  & Original       & 68                      & \textbf{54}               \\
                                         &                       & 3 Months later & \textbf{70}             & 48                        \\ \midrule
        \multirow{2}{*}{NeRF}            & \multirow{2}{*}{Math} & Original       & \textbf{97}             & \textbf{75}               \\
                                         &                       & 2 Months later & 91                      & 51                        \\ \midrule
        \multirow{2}{*}{Mesh}            & \multirow{2}{*}{Math} & Original       & \textbf{85}             & \textbf{56}               \\
                                         &                       & 2 Months later & 85                      & 49                        \\  
        \bottomrule
        \end{tabular}
\end{table*}

When using a prior map for localization, one would expect that changes to the scene would affect algorithm performance. In indoor environments, furniture and decorations are often reconfigured. Meanwhile, outdoor environments are affected by seasonal and lighting changes, while dynamic objects such as parked cars or bicycles may be added or removed. Our method must be resilient to such changes.

To demonstrate the robustness of our approach to scene change, we tested our camera localization system with data collected 2-3 months after the initial construction of the maps. The right-hand side of \figref{fig:changes-in-scene} shows two images from \texttt{MATH} with changes (shown with red circles) in the vegetation color, the location of parked cars, and the configuration of the bicycle stand. On the left-hand side, images from \texttt{ORI} show rearranged furniture and equipment.

In both locations, lighting variation between the 2-3 month period was clearly observed. The detailed results of localization performance are presented in \tabref{tab:change_in_env}. All three map representations show a small drop in retrieval and location rates. However, they can still localize around \SI{50}{\percent} of query images, demonstrating the robustness of matching against one unified synthetic image database, aided by learning-based visual feature detector SuperPoint and matcher SuperGlue.

\subsection{Ablation Study: Feature Detection and Matching}
\label{ablation_study2}
\begin{table*}[]
    \centering
    \caption{Average performance of different feature detector combinations for matching a set of ten real and synthetic image pairs. An example is shown in Fig.~\ref{fig:feature_detector_matcher}. Note that many of the features matched for Akaze and SIFT are incorrect.}
    \begin{tabular}{c|r|r}
         \toprule
         \textbf{Method} & \textbf{Feature Detected} & \textbf{Matched}   \\
         \midrule
         Akaze \cite{alcantarilla2013fast} + NN  & 410 &  18 \\
         SIFT \cite{lowe2004sift} + KNN & 841 & 21 \\
         R2D2 \cite{revaud2019r2d2} + NN  & 1024 & 69 \\
         D2Net \cite{Dusmanu2019D2net} + NN & 2932 & 132 \\
         SuperPoint \cite{SuperPoint2018} + SuperGlue \cite{SuperGlue2020} & 1024  & 201\\
         \bottomrule
    \end{tabular}
    \label{tab:feature_detector_comparison}
\end{table*}

\begin{figure}
    \centering
    \begin{subfigure}{0.9\linewidth}
        \includegraphics[width=\linewidth]{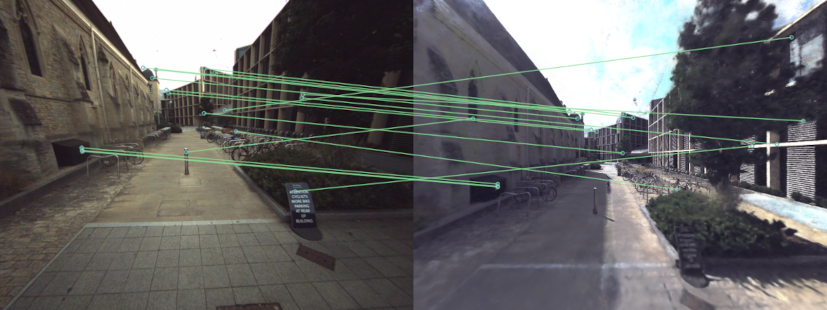}
        \caption{Detector: OpenCV Akaze, Matcher: Nearest Neighbor}
        \label{fig:akaze}
    \end{subfigure}
    \\[10pt]
    \begin{subfigure}{0.9\linewidth}
        \includegraphics[width=\linewidth]{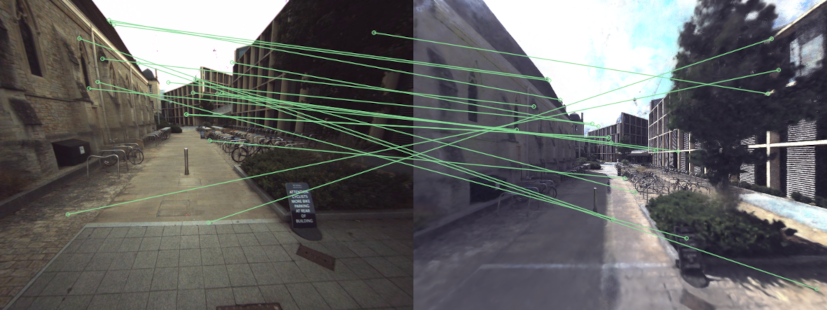}
        \caption{Detector: OpenCV SIFT, Matcher: Flann KNN}
        \label{fig:sift}
    \end{subfigure}
    \\[10pt]
    \begin{subfigure}{0.9\linewidth}
        \includegraphics[width=\linewidth]{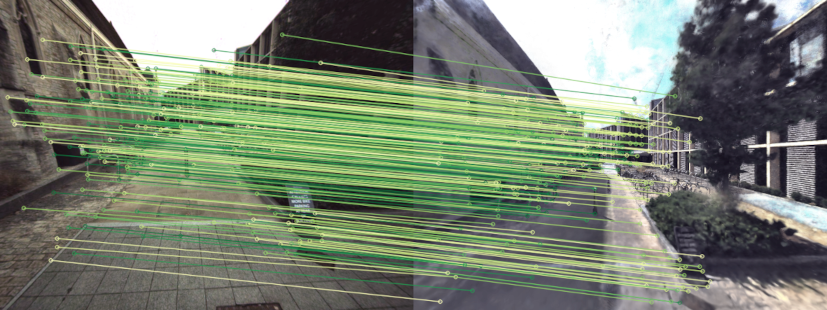}
        \caption{Detector: SuperPoint, Matcher: SuperGlue}
        \label{fig:SP_SG}
    \end{subfigure}

    \caption{Example of matching a camera image to a NeRF rendered image with different feature detectors.}
    \label{fig:feature_detector_matcher}
\end{figure}

As shown in Fig.~\ref{fig:render_image_comparison}, there is a significant reality gap between images captured by real cameras and synthetically rendered images. Each map representation has its own limitations when rendering the synthetic images. Images rendered from TLS point clouds often lack photorealism, exhibit a surreal quality, and have visible gaps corresponding to unscanned regions. Conversely, mesh-rendered images can contain areas with fragmented reconstruction and struggle to faithfully represent small and intricate geometric structures. For the NeRF-generated images, when the rendering poses differ largely from the training data viewpoints, one can often see a fog-like effect in the generated images. Perhaps learning-based feature detectors and matching techniques could help to bridge the domain gap between synthetic and real images and alleviate these issues.

In this section, we conduct a comparative analysis of different feature detectors and matchers using a representative dataset of ten image pairs drawn from across the three datasets. One image is a live camera image, and the other is a rendered image from either the point cloud, mesh, or NeRF pipelines. Fig.~\ref{fig:feature_detector_matcher} shows one scene with common challenges: vegetation, low-textured ground, and thin bike rails. \figref{fig:SP_SG} shows how learning-based approaches such as SuperPoint and SuperGlue can effectively identify and match numerous correct features. In contrast, traditional feature detectors like Akaze (\figref{fig:akaze}) and SIFT (\figref{fig:sift}) perform poorly in this scenario.

Tab.~\ref{tab:feature_detector_comparison} presents quantitative results for this test, comparing several learning-based and classic feature detection and matching techniques. Among the learning-based feature detectors, SuperPoint, R2D2, and D2Net show promising performance by matching many features. However, traditional methods like SIFT and Akaze match fewer features, many of which are incorrect. Note that R2D2 produces descriptors of size 128, and D2Net generates descriptors of size 512. Given that the SuperGlue matcher is pre-trained with descriptors of size 256, we use the Nearest Neighbour (NN) matcher with a ratio threshold test for R2D2 and D2Net. SuperPoint and R2D2 are capped at 1024 features, while D2Net, a dense image feature extractor, naturally yields more feature points.

In summary, leveraging learning-based feature detectors and matching techniques is crucial to mitigate the domain gap between camera-captured and synthetic images, enabling robust feature detection and matching across diverse image representations.

\section{Conclusions and Future Work}

\subsection{Conclusions}

Using readily available 3D color maps generated from SLAM missions or TLS scanners, we can repurpose these maps for localization tasks. In this paper, we introduce a localization system specifically designed to localize a single camera image within 3D color maps by sampling and generating a synthetic image database. We first demonstrated a pipeline to construct 3D prior maps using three distinct representations: point clouds, meshes, and NeRF. Each representation is then used to synthesize RGB and depth images. Following this, we proposed a strategy to define the set of rendering poses that optimize the creation of a representative 3D map while retaining a minimal set of database images.
The visual localization pipeline can estimate the camera pose of a query image through a retrieval and matching process, leveraging learning-based descriptors and feature detectors. We conducted a comprehensive analysis of localization performance across these representations and discussed their respective merits. Additionally, we offered a benchmark comparison with two purely vision-based localization systems to situate our results within the wider field of visual localization. Notably, both point cloud and mesh representations achieve a localization accuracy of \SI{55}{\percent} for query images, while the NeRF representation surpasses these, achieving a localization rate of \SI{72}{\percent}.

\subsection{Limitation and Future Work}

\subsubsection{Scene Change}
While we have studied the effect of scene changes within the context of a 3D color map, we acknowledge that there would likely be a performance decrease during gradual scene transitions over a period of months and years. In future research, we aim to devise a detection algorithm capable of promptly identifying these changes in real time, thus facilitating the identification of out-of-date regions of the map. Additionally, in scenarios where the device is equipped with a lidar sensor, we propose to update the existing map following a remapping process to ensure its continued accuracy and relevance.

\subsubsection{Synthetic Image}
All three map representations encounter specific challenges in accurately rendering scenes with minimal texture details. This difficulty is particularly notable in 3D mapping and reconstruction, where effectively capturing the RGB texture details of ground and vegetation proves challenging. Furthermore, during the online localization of our system, images that predominantly contain low-texture objects often yield insufficient features for matching, resulting in inaccurate pose estimations. To address this, we could prune the database images during online localization to retain only the most useful images and synthesizing new ones on-the-fly as needed.

\bibliography{library}

\end{document}